\pdfoutput=1
\documentclass[conference]{IEEEtran}
\IEEEoverridecommandlockouts
\usepackage{cite}
\usepackage{amsmath,amssymb,amsfonts}
\usepackage{amsthm}
\usepackage{algorithmic}
\usepackage{graphicx}
\usepackage{textcomp}
\usepackage{xcolor}

\newtheorem{key question}{\bf Key Question}
\newtheorem{corollary}{\bf Corollary}
\newtheorem{example}{\bf Example}
\newtheorem{definition}{\bf Definition}

\newtheorem{proposition}{\bf Proposition}
\newtheorem{lemma}{\bf Lemma}

\newtheorem{thm}{\bf Theorem}

\usepackage{cases}
\usepackage{subfigure}
\newtheorem{problem}{\bf Problem}
\usepackage{dsfont}
\usepackage{multirow}
\usepackage{array} 
\usepackage{modroman}
\DeclareMathOperator*{\argmax}{argmax}
\usepackage{graphics}
\usepackage{epstopdf}
\usepackage{tcolorbox}
\usepackage[linesnumbered,ruled,vlined]{algorithm2e}

\def\BibTeX{{\rm B\kern-.05em{\sc i\kern-.025em b}\kern-.08em
    T\kern-.1667em\lower.7ex\hbox{E}\kern-.125emX}}
    
\begin{document}

\title{Strategic Prompt Pricing for AIGC Services: A User-Centric Approach}

\author{Xiang Li, Bing Luo, Jianwei Huang, Yuan Luo\thanks{Xiang Li is with the Shenzhen Institute of Artificial Intelligence and Robotics for Society (AIRS), and the School of Science and Engineering (SSE), The Chinese University of Hong Kong, Shenzhen (E-mail: xiangli2@link.cuhk.edu.cn). Bing Luo is with Data Science Research Center, Duke Kunshan University, Jiangsu (E-mail: bl291@duke.edu). Jianwei Huang is with SSE, AIRS, Shenzhen Key Laboratory of Crowd Intelligence Empowered Low-Carbon Energy Network, and CSIJRI Joint Research Centre on Smart Energy Storage, The Chinese University of Hong Kong, Shenzhen, Guangdong, 518172, P.R. China (Corresponding Author, E-mail: jianweihuang@cuhk.edu.cn). Yuan Luo is with SSE, AIRS, and Shenzhen Key Laboratory of Crowd Intelligence Empowered Low-Carbon Energy Network, The Chinese University of Hong Kong, Shenzhen (E-mail: luoyuan@cuhk.edu.cn).}}

\maketitle

\begin{abstract}
The rapid growth of AI-generated content (AIGC) services has created an urgent need for effective prompt pricing strategies, yet current approaches overlook users' strategic two-step decision-making process in selecting and utilizing generative AI models. This oversight creates two key technical challenges: quantifying the relationship between user prompt capabilities and generation outcomes, and optimizing platform payoff while accounting for heterogeneous user behaviors. We address these challenges by introducing prompt ambiguity, a theoretical framework that captures users' varying abilities in prompt engineering, and developing an Optimal Prompt Pricing (OPP) algorithm. Our analysis reveals a counterintuitive insight: users with higher prompt ambiguity (i.e., lower capability) exhibit non-monotonic prompt usage patterns, first increasing then decreasing with ambiguity levels, reflecting complex changes in marginal utility. Experimental evaluation using a character-level GPT-like model demonstrates that our OPP algorithm achieves up to $31.72\%$ improvement in platform payoff compared to existing pricing mechanisms, validating the importance of user-centric prompt pricing in AIGC services.
\end{abstract}

\section{Introduction}
\begin{figure}[ht]
    \centering
    \includegraphics[width=0.48\textwidth]{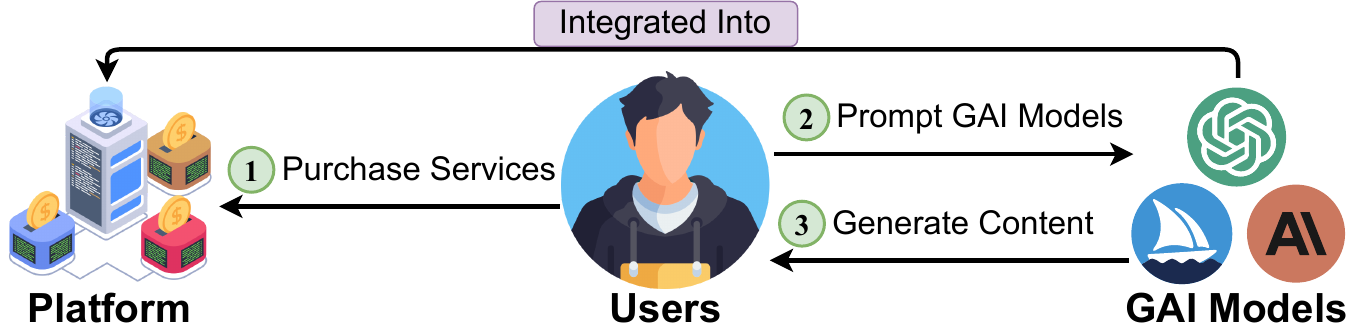}
    \vspace{-2pt}
    \caption{Framework of AIGC Service Process.}
    \label{Framework of AIGC Services}
    \vspace{-10pt}
\end{figure}

With impressive progress in generative artificial intelligence (GAI) \cite{Karapantelakis2024-ce}, AI-generated content (AIGC) services are growing in popularity, evidenced by platforms such as Monica \cite{Monica} and Poe \cite{gb}. Figure \ref{Framework of AIGC Services} illustrates a typical framework of the AIGC service process, where the platform usually integrates multiple GAI models, e.g., GPT-4 \cite{openai2024gpt4technicalreport} and GPT-4o \cite{openai2023gpt4o}, each with different performance of generating content in response to user prompts. Users will purchase AIGC services from the platform under a usage metric, i.e., the number of prompts \cite{cao2023comprehensivesurveyaigeneratedcontent}, to interact with distinct models and obtain desired content.

This novel AI service paradigm has promoted the emergence of related research \cite{10445209,10.1145/3589334.3645511,xu2023sparksgptsedgeintelligence,10504615,10172151,10388354}, as summarized in surveys \cite{wu2023aigeneratedcontentaigcsurvey,zhang2023completesurveygenerativeai,10221755}. For example, Du \emph{et al.} in \cite{10445209} designed a dynamic service selection scheme to enhance AIGC quality. D\"{u}tting \emph{et al.} in \cite{10.1145/3589334.3645511} proposed an AIGC advertising auction mechanism. Liu \emph{et al.} in \cite{10388354} incorporated semantic communication to achieve efficient content generation. However, these studies primarily address AIGC service optimization, leaving a significant research gap in platform prompt pricing design.

Current prompt pricing (e.g., \cite{Monica,gb,Bearly}) typically relies on GAI models' cost or performance metric, neglecting users' strategic decision-making process. In AIGC services, the platform decides the prompt price of each model and necessitates users to make two-step utilization decisions: which model to utilize and how to utilize it (i.e., the number of prompts for interaction \cite{cao2023comprehensivesurveyaigeneratedcontent,10221755}). Users will decide both the optimal model and prompt number to maximize individual payoff, based on their varying prompt capabilities \cite{sahoo2024systematicsurveypromptengineering}. Failing to account for such heterogeneous decisions in prompt pricing could not only reduce the platform's payoff, but also discourage users (e.g., those with low prompt capability) from AIGC services. Thus, to design the optimal prompt pricing mechanism, we first delve into users' strategic decisions in AIGC services, which motivates our key question as follows:

\begin{key question}
\label{q1}
What is the optimal utilization strategy for users with different prompt capabilities in AIGC services?
\end{key question}
The prompt-based content generation of AIGC services poses unique challenges to user decision-making. Specifically, the user's utility derived from generated content depends on both model performance and user prompts (i.e., prompt number and quality, as highlighted in \cite{sahoo2024systematicsurveypromptengineering,NEURIPS2022_9d560961}). This correlation and inherent uncertainty \cite{dong2024surveyincontextlearning} make it difficult to assess each prompt's value across diverse models and users, complicating decisions on selecting and prompting GAI models.

To address the above challenges from interactions between models and prompts, we introduce the concept of prompt ambiguity, inspired by in-context learning \cite{zhou2024mysteryincontextlearningcomprehensive,xie2022explanationincontextlearningimplicit,jiang2023latentspacetheoryemergent}. This mathematical framework quantifies user prompts as the probability of correctly conveying the intended task, enabling us to derive optimal utilization decisions for heterogeneous users. With users' optimal strategy in AIGC services, we can study our second key question of this paper:

\begin{key question}
\label{q2}
What is the platform’s optimal prompt pricing mechanism for AIGC services?
\end{key question}
The design of optimal prompt pricing presents a challenging bi-level non-convex optimization problem, due to the heterogeneity of users' prompt capabilities and strategic decisions. The platform must maximize payoff while accounting for users' two-step decision-making process, forming a complex bi-level structure. Prompt prices across various GAI models are also ineluctably intertwined with the trade-off between user strategies in selecting and prompting the model, resulting in the non-convexity of this mechanism design problem.

Against this background, we first transform the problem into an unconstrained piecewise form to determine optimal prompt pricing for users with consistent prompt capabilities (homogeneous case). Then, by exploring the relationship between user utilization decisions and model prompt prices in this simplified scenario, we establish the price upper bound for the more general heterogeneous user case. According to these insights, we propose an \underline{O}ptimal \underline{P}rompt \underline{P}ricing (OPP) algorithm, which decomposes prompt pricing design into tractable sub-problems for platform payoff maximization.

The key results and contributions of the paper are as follows:
\begin{itemize}
    \item \emph{Optimal Prompt Pricing Framework:} We address the fundamental challenge of modeling strategic interactions between the platform and heterogeneous users in AIGC services. Our framework transforms this bi-level optimization problem into tractable sub-problems through theoretical bounds analysis, providing the first systematic solution for prompt-based pricing mechanisms.
    \item \emph{User Utilization Strategy Analysis:} We tackle the challenge of quantifying diverse user prompt capabilities in AIGC services. By introducing prompt ambiguity based on in-context learning \cite{zhou2024mysteryincontextlearningcomprehensive,xie2022explanationincontextlearningimplicit,jiang2023latentspacetheoryemergent}, we establish a mathematical framework that precisely captures how users select and interact with different GAI models.
    \item \emph{Impact of User Prompt Ambiguity:} We solve the non-intuitive relationship between user prompt capability and optimal prompt usage. Our analysis reveals that higher prompt ambiguity (lower capability) creates a non-monotonic pattern in prompt number: first increases and then decreases with ambiguity levels, corresponding to variations in user marginal utility of additional prompts.
    \item \emph{Experimental Evaluation:} We bridge the gap between theoretical analysis and practical implementation using a character-level GPT-like model \cite{radford2018improving} and synthetic language data set \cite{zhou2024mysteryincontextlearningcomprehensive,xie2022explanationincontextlearningimplicit,jiang2023latentspacetheoryemergent}. The extensive experiments validate our analysis and demonstrate that our proposed OPP algorithm can improve the platform's payoff by up to $31.72\%$, compared to existing pricing mechanisms.
\end{itemize}

The remainder of this paper is organized as follows. Section \ref{S3} presents the system model and problem formulation in AIGC services. Section \ref{S4} and Section \ref{S5} respectively derive heterogeneous users' optimal utilization strategies and the platform's optimal prompt pricing. We provide experimental evaluation in Section \ref{S6} and conclude the paper in Section \ref{S7}. Due to the page limit, we leave detailed proofs of our results and additional theoretical analysis in the online appendix \cite{Online_appendix}.

\section{System Model and Problem Formulation}
\label{S3}
In this section, we first introduce the AIGC service process, a two-stage Stackelberg game between users and the platform. Then, we elaborated on modeling heterogeneous users and the platform's optimal prompt pricing mechanism design problem.

\subsection{The AIGC Service Process}\label{S3-A}
We consider a typical two-stage process of AIGC services between heterogeneous users and a platform, outlined below:
\begin{itemize}
    \item \textbf{Stage 1:} \textit{Platform} publishes a set of GAI models, each with a specific prompt price and performance characteristic of generating content in response to user prompts.
    \item \textbf{Stage 2:} \textit{Users} make two-step utilization decisions for AIGC services, determining which GAI model to utilize and the number of prompts required for its utilization.
\end{itemize}
\begin{figure}[ht]
    \centering
    \includegraphics[width=0.42\textwidth]{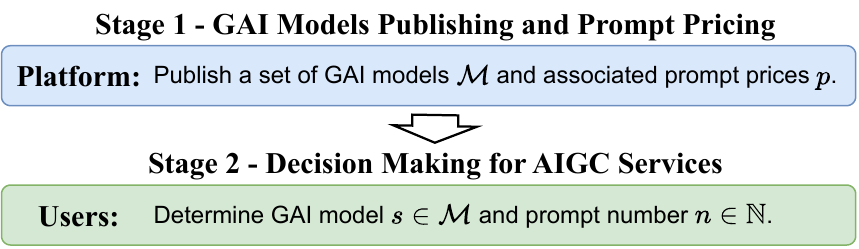}
    \caption{Two-stage Stackelberg Game of AIGC Services.}
    \label{AIGC Services Process}
\end{figure}

Next, we present each stage in detail, depicted in Figure \ref{AIGC Services Process}.
\subsubsection{Stage 1}
The platform offers a set of GAI models with different performances and decides each model's prompt price.
\begin{itemize}
    \item \textit{Published GAI Model Set:} The platform publishes a set of GAI models $\mathcal{M} = \{M_L, M_H\}$ with distinct performance tiers: a low-performance model $M_L$ and a high-performance model $M_H$. This binary categorization, prevalent in existing platforms (e.g., \cite{Monica,gb,Bearly}), enables focused theoretical analysis while maintaining generalizability to multi-model scenarios.
    \item \textit{Prompt Prices:} The platform establishes prompt prices $\boldsymbol{p} = \{p_m, \forall m \in \mathcal{M}\}$, where $p_m$ denotes the price per prompt for model $m\in \mathcal{M}$. This usage-based pricing scheme aligns with current industry practices. For example, Poe \cite{gb} charges approximately $\$0.008$ per prompt for GPT-4 \cite{openai2024gpt4technicalreport} and $\$0.01$ for GPT-4o \cite{openai2023gpt4o}.
    \item \textit{Operational Costs:} Each prompt incurs operational costs for the platform, represented by $\mathcal{C} = \{C_m, \forall m \in \mathcal{M}\}$, where $C_m$ denotes the per-prompt cost for GAI model $m\in \mathcal{M}$. These costs encompass both computation \cite{cao2023comprehensivesurveyaigeneratedcontent} and communication \cite{10221755} resources. For reference, GPT-4's operational cost is about $\$0.0036$ per prompt \cite{Patel2023-sp}.
\end{itemize}

\subsubsection{Stage 2} Users decide both which GAI model to utilize and the number of prompts required for its utilization.
\begin{itemize}
    \item \textit{Model Selection:} Users choose a specific model $s \in \mathcal{M}$ from the available options based on the platform's published model set $\mathcal{M}$ and pricing strategy $\boldsymbol{p}$.
    \item \textit{Prompt Strategy:} Users determine the optimal number of prompts $n \in \mathbb{N}$ needed for their task. This decision depends on both the selected model's performance and the user's prompt capability.
\end{itemize}
    
These decisions in Stage 2 are interdependent, and their optimal values are analyzed in detail in the following section.

\subsection{Modeling of Heterogeneous Users}\label{sys2.1}
This subsection characterizes the user's prompt behaviors, expected utility, and expected payoff in AIGC services.
\subsubsection{User's Prompt Behaviors}
\begin{figure}[ht]
    \centering
    \includegraphics[width=0.42\textwidth]{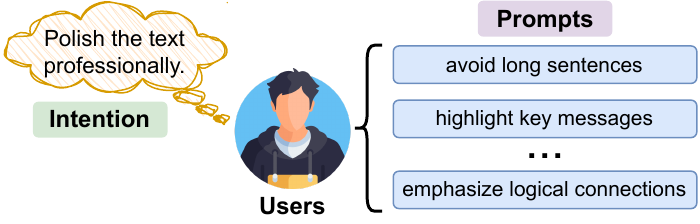}
    \caption{Example of User Prompts and Intention.}
    \label{Prompt Intention}
\end{figure}

Users interact with GAI models by conveying a specific intention (task) through prompts to generate desired content, such as instructions like ``avoid long sentences'' or ``highlight key messages'' when polishing text (see Figure \ref{Prompt Intention}). To formalize this interaction, we establish a mathematical framework where $\Gamma$ denotes a discrete and complete space of user intentions, with each intention $\gamma \in \Gamma$ being unique, and $\mathcal{X}_\gamma$ represents the set of user prompts associated with intention $\gamma$. Each prompt $x \in \mathcal{X}_\gamma$ is independently\footnote{This initial work focuses on scenarios where user prompts are independent of each other. For dependent cases, e.g., chain-of-thought prompting \cite{NEURIPS2022_9d560961}, the analysis is similar but involves more complicated prompt correlations.} generated from a distribution \cite{PIERACCINI1992283,hidden} conditioned on the corresponding intention $\gamma$, providing a foundational structure for analyzing user-model interactions in AIGC services.

In practice, users' varying prompt capabilities often result in ambiguous communication of their target intention $\gamma \in \Gamma$. Drawing from in-context learning literature \cite{zhou2024mysteryincontextlearningcomprehensive,xie2022explanationincontextlearningimplicit,jiang2023latentspacetheoryemergent}, we introduce the concept of prompt ambiguity to quantify these heterogeneous prompt capabilities:

\begin{definition}[Prompt Ambiguity]
\label{Def_ambiguity}
A user prompt $x\in\mathcal{X}_\gamma$ is $\epsilon$-ambiguous\footnote{To avoid trivial cases, we assume $\epsilon\in(0,1)$. If $\epsilon=0$, the user always requires only one prompt. Conversely, when $\epsilon=1$, the prompt is meaningless.} for intention $\gamma\in\Gamma$ with $\epsilon\in(0,1)$, if
\begin{align}\label{eq1}
    \textrm{Prob}\left(\gamma|x\right)\geq 1-\epsilon,
\end{align}
which indicates that the probability of correctly inferring the target intention $\gamma$ from user prompt $x$ is at least $1-\epsilon$.
\end{definition}
Building upon Definition \ref{Def_ambiguity} and existing literature \cite{zhou2024mysteryincontextlearningcomprehensive,xie2022explanationincontextlearningimplicit,jiang2023latentspacetheoryemergent}, we establish the relationship between user prompts and target intention in Lemma \ref{Lemma1}.
\begin{lemma}\label{Lemma1}
    For a user with prompt ambiguity $\epsilon$, the probability of inferring target intention $\gamma \in \Gamma$ from $k \in \mathbb{N}^+$ prompts is:
    \begin{align}\label{eq3}
    \text{Prob}(\gamma_1 = \gamma,\cdots,\gamma_k = \gamma|x_1,\cdots,x_k) \geq 1-\epsilon^k,
\end{align}
where $x_k$ denotes the user's $k$-th prompt and $\gamma_k$ represents the intention conveyed by prompt $x_k$, for any $k \in \mathbb{N}^+$.
\end{lemma}
The detailed proof of Lemma \ref{Lemma1} is provided in the appendix \cite{Online_appendix}. To capture the diversity in users' prompt capabilities, we employ a probability density function $f(\epsilon)$ to characterize the distribution of prompt ambiguity across the user population.
\subsubsection{User's Expected Utility} Within intention space $\Gamma$, a user's utility from generated content is determined by two factors: the selected model's performance and the probability of successfully conveying the target intention through prompts. For any published model $m \in \mathcal{M}$, we define $U_m$ as the utility that maps from intention space to generated content, indicating the model's performance in responding to user intentions (e.g., prediction accuracy in semantic tasks \cite{zhou2024mysteryincontextlearningcomprehensive}). Accordingly, the utility mapping of the user's selected model $s\in\mathcal{M}$ is $U_s$. Based on Lemma \ref{Lemma1}, for a user's GAI model selection $s \in \mathcal{M}$, prompt number $n \in \mathbb{N}$, and prompt ambiguity $\epsilon \in (0,1)$, the user's expected utility is expressed as:
\begin{align}\label{eq_userutility}
    \left(1-\epsilon^n\right)\cdot U_s,
\end{align}
where $U_s$ represents the utility of model $s\in\mathcal{M}$.
\subsubsection{User's Expected Payoff}
The user's payoff $\pi_{\text{user}}$ is defined as the difference between expected utility $(1-\epsilon^n)U_s$ and the total prompting price $np_s$:
\begin{align}\label{eq_userpayoff}
    \pi_{\textrm{user}}(s,n,\boldsymbol{p},\epsilon)=\left(1-\epsilon^n\right)\cdot U_s-n\cdot p_s.
\end{align}

Users make optimal decisions by selecting model choice $s^* \in \mathcal{M}$ and prompt number $n^* \in \mathbb{N}$ to maximize payoff $\pi_{\text{user}}$ in \eqref{eq_userpayoff}. These decisions are based on prompt prices $\boldsymbol{p}$ and their individual prompt ambiguity $\epsilon \in (0,1)$. To emphasize the interdependence between optimal decisions, we denote them as $s^*(n^*,\boldsymbol{p})$ and $n^*(s,p_s,\epsilon)$, respectively. This formalization provides the foundation for analyzing the platform's prompt pricing mechanism design problem.

\subsection{Platform's Prompt Pricing Mechanism Design Problem}
In AIGC services, the platform decides the optimal prompt pricing $\boldsymbol{p}^*=\{p_m^*,\forall m\in\mathcal{M}\}$ to maximize its payoff across all GAI models and users. We first summarize the platform's expected payoff $\pi_{\textrm{platform}}(\boldsymbol{p})$ and then formally introduce its problem of prompt pricing mechanism design.

For the platform, its expected payoff $\pi_{\textrm{platform}}(\boldsymbol{p})$ is the aggregate of that from each GAI model. Specifically, given any model $m\in\mathcal{M}$, the platform's per-prompt payoff is $p_m-C_m$, where $p_m$ and $C_m$, respectively, represent the model's prompt price and prompt cost. Through the prompt ambiguity density function $f(\epsilon)$, we can further determine the expected number of user prompts for model $m\in\mathcal{M}$:
\begin{align}
    N_m(\boldsymbol{p})=\int_0^1 f(\epsilon)\cdot n^*(m,p_{m},\epsilon)\cdot\mathds{1}_{s^*(n^*,\boldsymbol{p})=m}\,d\epsilon,
\end{align}
where indicator function $\mathds{1}_{s^*(n^*,\boldsymbol{p})=m}$ indicates that the optimal GAI model for the user is $m\in\mathcal{M}$. Accordingly, the expected payoff of the platform in AIGC services is:
\begin{align}\label{eq5}
    \pi_{\textrm{platform}}(\boldsymbol{p})=&\sum\nolimits_{m\in\mathcal{M}} \left(p_m-C_m\right)\cdot N_m(\boldsymbol{p}).
\end{align}

To maximize payoff $\pi_{\textrm{platform}}(\boldsymbol{p})$ in \eqref{eq5}, the platform needs to account for each user's self-interested decisions $s^*(n^*,\boldsymbol{p})$ and $n^*(s,p_s,\epsilon)$. By incorporating heterogeneous users' strategic decisions, we formulate the platform's optimal prompt pricing problem as a bi-level optimization in Problem \ref{Pb2}.
\begin{tcolorbox}
\begin{problem}[Prompt Pricing Mechanism Design]\label{Pb2}
    \begin{align*}
    \max_{\boldsymbol{p}}&\textrm{ }\sum\nolimits_{m\in\mathcal{M}} \left(p_m-C_m\right)\cdot N_m(\boldsymbol{p}),\\
    s.t.&\textrm{ }\pi_{\textrm{user}}(s,n,\boldsymbol{p},\epsilon)=\left(1-\epsilon^n\right)\cdot U_s-n\cdot p_s,\\
    &\hspace{-1pt}N_m(\boldsymbol{p})=\int_0^1 f(\epsilon)\cdot n^*(m,p_{m},\epsilon)\cdot\mathds{1}_{s^*(n^*,\boldsymbol{p})=m}\,d\epsilon,\\
    &\textrm{ }\hspace{-3pt}n^*(s,p_s,\epsilon)=\argmax_{n\in\mathbb{N}} \textrm{ }\pi_{\textrm{user}}(s,n,\boldsymbol{p},\epsilon),\textrm{ } \forall s\in\mathcal{M},\\
    &\textrm{ }\hspace{-3pt}s^*(n^*,\boldsymbol{p})=\argmax_{s\in\mathcal{M}}\textrm{ }\pi_{\textrm{user}}(s,n^*(s,p_s,\epsilon),\boldsymbol{p},\epsilon),\\
    var.&\textrm{ }\boldsymbol{p}=\{p_m,\forall m\in\mathcal{M}\}.
\end{align*}
\end{problem}
\end{tcolorbox}
Having established the system model and problem formulation for AIGC services, we tackle this two-stage Stackelberg game using backward induction \cite{Mas-Colell1995-si}. Our study follows a sequential approach: Section \ref{S4} examines users' strategic utilization decisions in Stage 2, and subsequently, Section \ref{S5} leverages these insights to derive the platform's optimal prompt pricing mechanism in Stage 1.

\section{Heterogeneous Users' Utilization Strategy}\label{S4}
This section analyzes heterogeneous users' optimal utilization strategies in AIGC services. Our analysis progresses from single-user optimization to multiple-user scenarios, revealing how prompt ambiguity and pricing affect users' strategic behaviors and expected payoff.

\subsection{The User's Optimal Utilization Strategy}
According to the published model set $\mathcal{M}$ and prompt pricing $\boldsymbol{p}=\{p_m,\forall m\in\mathcal{M}\}$, each user maximizes individual payoff $\pi_{\textrm{user}}(s,n,\boldsymbol{p},\epsilon)$ by deciding the optimal model $s^*(n^*,\boldsymbol{p})$ and prompt number $n^*(s,p_s,\epsilon)$. Given model decision $s\in\mathcal{M}$ and ambiguity $\epsilon\in(0,1)$, Theorem \ref{T1} derives the optimal prompt strategy $n^*(s,p_s,\epsilon)$ for the user.
\begin{thm}
    \label{T1}
    For any selected model $s\in\mathcal{M}$ with utility mapping $U_s$ and prompt price $p_s$, a user with prompt ambiguity $\epsilon \in (0, 1)$ has the following optimal strategy:
        \begin{subnumcases}{\label{eqT1}n^*(s,p_s,\epsilon)=}
    \hspace{-3pt}\lfloor \log_{\epsilon}\frac{\epsilon p_s}{(1-\epsilon)U_s} \rfloor, \hspace{-9pt}&if $0\leq p_s\leq(1-\epsilon)U_s,$\\
    \hspace{-3pt}0, &if $p_s>(1-\epsilon)U_s$,
\end{subnumcases}
where $\lfloor\cdot\rfloor$ is the floor function.
\end{thm}
We give the proof of Theorem \ref{T1} in online appendix \cite{Online_appendix}. Theorem \ref{T1} provides a theoretical foundation for understanding the relationship between optimal prompt number $n^*(s,p_s,\epsilon)$ and two key factors: prompt price $p_s$ and utility mapping $U_s$. For any selected model $s\in\mathcal{M}$, utility mapping $U_s$ establishes a price ceiling of $(1-\epsilon)U_s$ that a user with ambiguity $\epsilon$ will accept. The optimal number of prompts will increase with $U_s$ and decrease with prompt price $p_s$.

Following Theorem \ref{T1}, we can conclude the user's optimal model selection $s^*(n^*,\boldsymbol{p})$ from model set $\mathcal{M}$:
\begin{align}\label{eq7}
    s^*(n^*,\boldsymbol{p})=\argmax\nolimits_{s\in\mathcal{M}}&(1-\epsilon\char94{\lfloor \log_{\epsilon}\frac{\epsilon p_s}{(1-\epsilon)U_s} \rfloor})\cdot U_s\nonumber\\&-\lfloor \log_{\epsilon}\frac{\epsilon p_s}{(1-\epsilon)U_s} \rfloor \cdot p_s.
\end{align}

This formulation captures the balance between utility maximization and price minimization in model selection.
\subsection{Multiple Users with Heterogeneous Prompt Ambiguities}
After studying single user's utilization decisions $s^*(n^*,\boldsymbol{p})$ and $n^*(s,p_s,\epsilon)$, we proceed to extend our analysis to encompass multiple users with heterogeneous prompt ambiguities. In particular, we start from the upper bound of heterogeneous users' optimal prompt number $\bar{n}^*(m)\in\mathbb{N}^+$ in Lemma \ref{Lemma2}, in order to investigate the impact of ambiguity $\epsilon$ on users' optimal strategies. This will form the foundation for designing the optimal prompt pricing mechanism in Stage 1.

\begin{lemma}\label{Lemma2}
    For users with varying ambiguity $\epsilon\in(0,1)$, the upper bound $\bar{n}^*(m)$ on users' optimal prompt number to GAI model $m\in\mathcal{M}$ is:
    \begin{align}\label{eq8}
        \bar{n}^*(m)=\min\left\{k\in\mathbb{N}^+:\frac{k^k}{(k+1)^{k+1}}<\frac{p_m}{U_m}\right\}.
    \end{align}
\end{lemma}
Lemma \ref{Lemma2} identifies the critical ratio ${p_m}/{U_m}$ that affects the upper bound $\bar{n}^*(m)$ of heterogeneous users' optimal prompt number for model $m\in\mathcal{M}$. Building on this, Proposition \ref{Pro1} elaborates on the relationship between prompt ambiguity $\epsilon\in(0,1)$ and the optimal prompt strategy $n^*(s,p_s,\epsilon)$ of users.
\begin{proposition}\label{Pro1}
    Given model decision $s\in\mathcal{M}$, utility mapping $U_s$ and prompt price $p_s$, users' optimal prompt strategy $n^*(s,p_s,\epsilon)$ varies with ambiguity $\epsilon\in(0,1)$ as follows:
    \begin{itemize}
        \item if $p_s=0$: $n^*(s,p_s,\epsilon)=\infty$.
        \item if $p_s\in(0,U_s/4]$: $n^*(s,p_s,\epsilon)$ first increases  and then decreases with $\epsilon$.
        \item if $p_s\in(U_s/4,U_s)$: $n^*(s,p_s,\epsilon)$ decreases with $\epsilon$.
        \item if $p_s\in[U_s,\infty)$: $n^*(s,p_s,\epsilon)=0$.
    \end{itemize}
\end{proposition}

We give the proof of Proposition \ref{Pro1} in appendix \cite{Online_appendix}. Proposition \ref{Pro1} reveals that users' optimal prompt strategy exhibits non-monotonic variation with ambiguity $\epsilon$. The behavior can be categorized into three distinct cases, depicted in Figure \ref{Vary Epsilon}:
\begin{itemize}
    \item \textbf{Extreme Cases:} When $p_s = 0$ or $p_s \geq U_s$, users either continuously prompt the model $s\in\mathcal{M}$ or completely avoid AIGC services of model $s\in\mathcal{M}$.\begin{figure}[ht]
    \centering
    \includegraphics[width=0.44\textwidth]{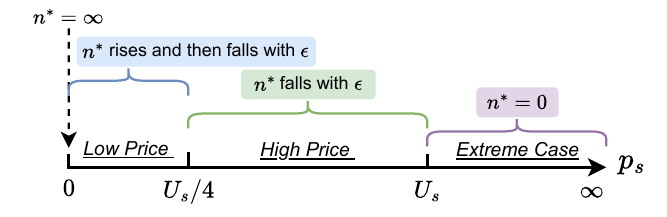}
    \caption{The Optimal Prompt Number $n^*(s,p_s,\epsilon)$ of Users with Varying Levels of Prompt Ambiguity $\epsilon\in(0,1)$.}
    \label{Vary Epsilon}
\end{figure}
    \item \textbf{Low Price Range} ($p_s \in (0,U_s/4]$): The optimal prompt number follows an inverse U-shaped pattern as ambiguity rises. Users with moderate prompt capability will use more prompts than those with very high or low capabilities, due to higher marginal utility of additional prompts.
    \item \textbf{High Price Range} ($p_s \in (U_s/4,U_s)$): Decreased prompt capability (higher ambiguity $\epsilon$) consistently reduces the optimal prompt number. The elevated price amplifies the disadvantage of lower prompt capability, discouraging users with higher prompt ambiguity from AIGC services.
\end{itemize}
\subsection{The User's Optimal Expected Payoff}
Given the prompt strategy $n^*(s,p_s,\epsilon)\in\mathbb{N}$ in Theorem \ref{T1} and GAI model decision $s^*(n^*,\boldsymbol{p})\in\mathcal{M}$ in \eqref{eq7},  we summarize the user's optimal payoff $\pi_{\textrm{user}}^*(s^*,n^*,\boldsymbol{p},\epsilon)$ as follows:
\begin{align}\label{eq9}
    \pi_{\textrm{user}}^*(s^*,n^*,\boldsymbol{p},\epsilon)=&\left(1-\epsilon^{n^*(s^*,p_{s^*},\epsilon)}\right) \cdot U_{s^*}\nonumber\\&-n^*(s^*,p_{s^*},\epsilon)\cdot p_{s^*},
\end{align}
where $U_{s^*}$ and $p_{s^*}$ are the utility mapping and prompt price of the user's optimal model $s^*(n^*,\boldsymbol{p})\in\mathcal{M}$, respectively.

With $\pi_{\textrm{user}}^*(s^*,n^*,\boldsymbol{p},\epsilon)$ in \eqref{eq9}, we aim to discuss the effect of prompt ambiguity $\epsilon\in(0,1)$ on the user's optimal expected payoff, as concluded in Proposition \ref{Ob1}.
\begin{proposition}\label{Ob1}
    The user's optimal payoff $\pi_{\textrm{user}}^*(s^*,n^*,\boldsymbol{p},\epsilon)$ decreases with prompt ambiguity $\epsilon\in(0,1)$.
\end{proposition}

Given the complex nature of users' two-step decisions (as illustrated in Proposition \ref{Pro1}), the proof of Proposition 2 requires careful analysis and is provided in detail in appendix \cite{Online_appendix}. This proposition indicates a crucial insight: while users may adapt their prompt strategy $n^*(s, p_s, \epsilon)$ in various ways depending on the ratio $p_s/U_s$ (particularly when $p_s/U_s \in (0, 1/4]$), their optimal payoff $\pi_{\textrm{user}}^*(s^*,n^*,\boldsymbol{p},\epsilon)$ invariably decreases as ambiguity $\epsilon$ increases. This demonstrates that despite users' strategic adjustments to maximize their payoff under different ambiguity levels, the negative impact of increased prompt ambiguity on the optimal payoff cannot be fully mitigated.

Having completed our analysis of heterogeneous users' optimal decisions and payoffs, we now proceed to examine the optimal prompt pricing in Stage 1 of AIGC services.
\section{Platform's Prompt Pricing Mechanism Design}\label{S5}
This section derives the platform's optimal prompt pricing mechanism by examining two distinct user scenarios. First, we focus on users with consistent prompt ambiguity, i.e., homogeneous cases, to establish a closed-form solution for optimal pricing. Then, based on these theoretical results, we extend our analysis to heterogeneous users and develop an efficient prompt pricing algorithm for platform payoff maximization.
\subsection{The Optimal Prompt Pricing for Homogeneous Users}
For a published GAI model set $\mathcal{M}$, the platform aims to determine the optimal prompt pricing $\boldsymbol{p}^*=\{p_m^*,\forall m\in\mathcal{M}\}$ that maximizes its payoff $\pi_{\textrm{platform}}(\boldsymbol{p})$ in \eqref{eq5}, formulated as a bi-level optimization in Problem \ref{Pb2}. Theorem \ref{T2} establishes the optimal prompt price for each model when serving homogeneous users with prompt ambiguity $\epsilon \in (0, 1)$.
\begin{thm}\label{T2}
For a model set $\mathcal{M}$ where each model $m\in\mathcal{M}$ has utility mapping $U_m$ and prompt cost $C_m$, the platform's optimal prompt pricing $\boldsymbol{p}^*=\{p_m^*,\forall m\in\mathcal{M}\}$ for homogeneous users with prompt ambiguity $\epsilon \in (0, 1)$ is:
    \begin{subnumcases}{p_m^*=}
        \epsilon^{\sigma(m,\epsilon)-1}(1-\epsilon)U_m, &if $m=m^*$,\label{eq10}\\
        U_m, &if $m\neq m^*$,
    \end{subnumcases}
    where $\sigma(m,\epsilon)=\min\{k\in\mathbb{N}:(k+1)\epsilon^{k}-k\epsilon^{k-1}<\frac{C_m}{(1-\epsilon)U_m}\}$ and $m^*=\argmax_{m\in\mathcal{M}} (\epsilon^{\sigma(m,\epsilon)-1}(1-\epsilon)U_m-C_m)\cdot \sigma(m,\epsilon)$.
\end{thm}

We give the proof of Theorem \ref{T2} in the online appendix \cite{Online_appendix}. Theorem \ref{T2} formalizes the optimal prompt price $p_m^*$ of each model $m\in\mathcal{M}$, which increases with utility mapping $U_m$ and decreases with prompt cost $C_m$.

\begin{figure}[ht]
    \centering
    \includegraphics[width=0.48\textwidth]{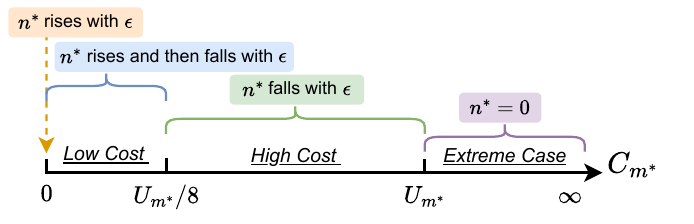}
    \caption{The Optimal Prompt Number $n^*(m^*,p_{m^*},\epsilon)$ of Homogeneous Users with Different Ambiguity $\epsilon$ under $p^*_{m^*}$ in \eqref{eq10}.}
    \label{Vary Epsilonopt}
\end{figure}

Building on Theorem \ref{T2}, and in preparation for designing prompt pricing with heterogeneous users, we further need to understand the impact of ambiguity $\epsilon$ on optimal pricing $\boldsymbol{p}^*$.
\begin{corollary}\label{Coro1}
For model $m^*\in\mathcal{M}$ with utility mapping $U_{m^*}$, prompt cost $C_{m^*}$, and optimal prompt price $p^*_{m^*}$, the user's optimal strategy $n^*(m^*, p^*_{m^*}, \epsilon)$ varies with prompt ambiguity $\epsilon \in (0, 1)$ as follows:
    \begin{itemize}
        \item if $C_{m^*}=0$: $n^*(m^*,p^*_{m^*},\epsilon)$ increases with $\epsilon$.
        \item if $C_{m^*}\in(0,U_{m^*}/8]$: $n^*(m^*,p^*_{m^*},\epsilon)$ first increases and then decreases with $\epsilon$.
        \item if $C_{m^*}\in(U_{m^*}/8,U_{m^*})$: $n^* (m^*,p^*_{m^*},\epsilon)$ decreases with $\epsilon$.
        \item if $C_{m^*}\in[U_{m^*},\infty)$: $n^*(m^*,p^*_{m^*},\epsilon)=0$.
    \end{itemize}
\end{corollary}
We give the proof of Corollary \ref{Coro1} in appendix \cite{Online_appendix}. This relationship between prompt cost and optimal strategy is illustrated in Figure \ref{Vary Epsilonopt}. In zero-cost scenario ($C_{m^*} = 0$), increasing ambiguity leads the platform to encourage more prompts to maximize payoff. For low cost ($C_{m^*} \in (0,U_{m^*}/8]$), the platform initially reduces prices as ambiguity grows to stimulate higher prompt numbers, but beyond a threshold, it raises prices, reducing prompt usage. With higher cost ($C_{m^*} \in (U_{m^*}/8,U_{m^*})$), increased ambiguity consistently leads to fewer prompts under optimal pricing, as incentivizing additional prompts becomes unprofitable.

Through Corollary \ref{Coro1}, Proposition \ref{Ob2} highlights the impact of prompt ambiguity $\epsilon\in(0,1)$ on the platform's optimal payoff, corresponding to that on users in Proposition \ref{Ob1}.
\begin{proposition}\label{Ob2}
    The platform's optimal payoff $\pi_{\textrm{platform}}^*(\boldsymbol{p}^*)$ under homogeneous users decreases with ambiguity $\epsilon\in(0,1)$.
\end{proposition}

The complete proof of Proposition \ref{Ob2} is provided in appendix \cite{Online_appendix}. This proposition implies that prompt ambiguity has a detrimental effect on both the platform's and users' payoffs. Notably, an increase in prompt ambiguity $\epsilon$ does not guarantee a higher number of user prompts, as shown in Proposition \ref{Pro1}. Even though the platform strategically optimizes the balance between prompt pricing and user prompt frequency, both parties experience diminishing optimal payoffs, creating a suboptimal outcome for all stakeholders.

\subsection{The Optimal Prompt Pricing for Heterogeneous Users}
Having analyzed the homogeneous case, we now address the more practical scenario of AIGC services with heterogeneous user prompt ambiguity. We begin by conducting a theoretical analysis of Problem \ref{Pb2} within this heterogeneous context, which is a multi-variable non-convex optimization (strongly NP-hard even in linear forms \cite{hansen}). Then, we exploit the interaction between user behavior, GAI models, and prompt pricing to design an efficient algorithm for solving it.
\subsubsection{Theoretical Analysis} 
Based on heterogeneous users' optimal prompt strategy in Theorem \ref{T1}, we first investigate the payoff maximization for the platform under any single model $m\in\mathcal{M}$. With Proposition \ref{Pro1} and Lemma \ref{Lemma2}, we reformulate such a problem into an unconstrained piecewise optimization in Problem \ref{Pb3}, where $\lambda_1(k)\in(0,1)$ and $\lambda_2(k)\in(0,1)$ are the implicit solutions of equation $\epsilon^{k-1}(1-\epsilon)U_m=p_m$ with $\lambda_1(k)\leq\lambda_2(k)$, and upper bound $\bar{n}^*(m)$ in \eqref{eq8}.
\begin{tcolorbox}
\begin{problem}[Prompt Pricing for Single GAI Model]\label{Pb3}
\begin{align*}
    \max_{p_m}(p_m-C_m)(\int_0^{1-\frac{p_m}{U_m}}f(\epsilon)d\epsilon+\sum_{k=2}^{\bar{n}^*}\int_{\lambda_1(k)}^{\lambda_2(k)}f(\epsilon)d\epsilon).
\end{align*}
\end{problem}
\end{tcolorbox}

This transformation significantly reduces the complexity of optimal prompt pricing for single model. We illustrate this with Example \ref{Example 1}, using a uniform distribution of prompt ambiguity.
\begin{example}[Uniform Distribution of User Prompt Ambiguity]\label{Example 1}
For prompt ambiguity uniformly distributed over $[\epsilon_\textrm{min},\epsilon_\textrm{max}]$ where $\epsilon_\textrm{min} < \epsilon_\textrm{max}$:
\begin{itemize}
    \item Problem \ref{Pb3} exhibits strict concavity on each piecewise segment $k\in[2,\bar{n}^*(m)]$.
    \item When $\epsilon_\textrm{min} = 0$ and $C_m > 0$, the optimal prompt price is $p^*_m = (U_m + C_m)/2$.
\end{itemize}
\end{example}

Example \ref{Example 1} reveals that under uniform ambiguity distribution, the platform prioritizes users with lower ambiguity to maximize $\pi_{\textrm{platform}}(\boldsymbol{p})$. When the ambiguity distribution's lower bound is zero, the optimal price $p^*_m$ depends solely on utility mapping $U_m$ and prompt cost $C_m$.

Following the prompt pricing for single GAI model, we extend our exploration to the entire model set $\mathcal{M}$ and incorporate users' optimal model decisions derived in \eqref{eq7}. \begin{algorithm}\caption{The Optimal Prompt Pricing (OPP).}\label{Algo1}
    \KwIn{prompt cost $C_L$ and $C_H$, utility mapping $U_L$ and $U_H$, user prompt ambiguity density function $f(\epsilon)$, and step size $\alpha$.}
    \KwOut{the optimal prompt pricing $\boldsymbol{p}^*=\{p_L^*,p_H^*\}$ and resultant payoff $\pi_\textrm{platform}^*$.}
    $\pi_\textrm{platform}^*=0$\;
    \For{$p_L=C_L:\alpha:U_L$}{Derive $\bar p_H(M_L,\epsilon)$ in \eqref{eq11}\;
Construct $\tilde{f}(\epsilon,p_H)$ in \eqref{eq13} and $G(p_H)$ in \eqref{eq14}\;
$\hat{p}_H=\{x\in[C_H,\bar{p}_H(M_L,0)]:\partial G(x)/\partial x=0\}$\;
$p_H=\argmax_{x\in\{C_H, \hat{p}_H, \bar p_H(M_L,0)\}}G(x)$\;
\If{$\pi_\textrm{platform}(p_L,p_H)\geq\pi^*_\textrm{platform}$}{$\boldsymbol{p}^*=\{p_L,p_H\}$\;
$\pi^*_\textrm{platform}=\pi_\textrm{platform}(p_L,p_H)$\;}
}
Return $\boldsymbol{p}^*$ and $\pi^*_\textrm{platform}$\;
\end{algorithm}Specifically, for any two models $m$ and $m'$ in $\mathcal{M}$, we establish upper bound $\bar{p}_m(m',\epsilon)$ on prompt price $p_m$ such that users with prompt ambiguity $\epsilon\in(0,1)$ will prefer model $m\in\mathcal{M}$ over $m'\in\mathcal{M}$, and vice versa, expressed as below:
\begin{align}\label{eq11}
    \bar{p}_m(m',\epsilon)=&\left[(1-\epsilon^{\tau(m,m',\epsilon)})\cdot U_m+n^*(m',p_{m'},\epsilon)\cdot p_{m'}\right.\nonumber\\&-\left.(1-\epsilon^{n^*(m',p_{m'},\epsilon)}) \cdot U_{m'}\right]\cdot\frac{1}{\tau(m,m',\epsilon)},
\end{align}
where $n^*(m',p_{m'},\epsilon)$ is in Theorem \ref{T1} and $\tau(m, m', \epsilon)$ is:
\begin{align}\label{eq12}
    \tau(m, m', \epsilon)=&\min\{k\in\mathbb{N}^+:(1-\epsilon^{n^*})\cdot U_{m'}-n^*\cdot p_{m'}\nonumber\\&<(1-\epsilon^k)\cdot U_m-k\epsilon^k(1-\epsilon)\cdot U_m\}.
\end{align}

From \eqref{eq11}, the optimal prompt pricing involves multi-level interactions with heterogeneous users' prompt strategy and model decision, especially given model set $\mathcal{M}$ and ambiguity distribution $f(\epsilon)$. This interrelation introduces non-convexity and significant computational complexity, making it difficult to derive a closed-form solution in general scenarios. Hence, we next aim to develop an efficient algorithm for Problem \ref{Pb2}.
\subsubsection{Algorithm Design}
To address Problem 1, we develop the \underline{O}ptimal \underline{P}rompt \underline{P}ricing (OPP) algorithm (Algorithm \ref{Algo1}) through systematic decomposition into tractable sub-problems. Our approach, inspired by finite element methods \cite{Bangerth2003-hy,Szabo2021-ml,Babuska2001-lv}, enhances computational efficiency by reducing the search space to a single dimension compared to exhaustive search. The algorithm integrates price upper bounds and strategically partitions the ambiguity distribution based on user decisions.

\textit{Workflow of Algorithm \ref{Algo1}:} Given prompt cost and utility mapping of GAI models in set $\mathcal{M}=\{M_L,M_H\}$, and user prompt ambiguity density function $f(\epsilon)$, we identify the upper bound $\bar p_H(M_L,\epsilon)$ in \eqref{eq11}, corresponding to prompt price $p_L$ of model $M_L$ (Line 2-3). Then, we partition $f(\epsilon)$ according to the optimal model decisions of heterogeneous users in AIGC services, formulated as a function of prompt price $p_H$:
\begin{subnumcases}{\tilde{f}(\epsilon,p_H)=\label{eq13}}
    f(\epsilon), &\textrm{ } if\textrm{ } $p_H\leq \bar p_H(M_L,\epsilon),$\\
    0, &\textrm{ } if\textrm{ } $p_H> \bar p_H(M_L,\epsilon)$.
\end{subnumcases}

Using revised ambiguity density function $\tilde{f}(\epsilon,p_H)$ from \eqref{eq13}, we construct $G(p_H)$ in \eqref{eq14}, aligning it with optimizing price $p_H$ for platform payoff maximization (Line 4):  
\begin{align}\label{eq14}
    G(p_H)=&\int_0^1 \tilde{f}(\epsilon,p_H)\cdot\left[(p_H-C_H)\cdot n^*(M_H,p_H,\epsilon)\right.\nonumber\\&\left.-(p_L-C_L)\cdot n^*(M_L,p_L,\epsilon)\right]d\epsilon.
\end{align}

Through the preceding transformation and simplification, we derive the optimal prompt price $p^*_H$ for any given $p_L$ of model $M_L$ (Line 5-6). By iteratively updating the prompt pricing mechanism, the OPP algorithm ultimately outputs the optimal one and resultant payoff for the platform (Lines 7-9). 

\section{Experimental Evaluation}
\label{S6}
This section outlines the experimental setup and presents results validating the theoretical analysis of user strategies and prompt pricing, highlighting their impact on platform payoff.
\subsection{Experimental Setup} 
We adopt a character-level GPT-like model \cite{radford2018improving}, with $21$ million parameters and a block size of $256$ to simulate practical AIGC services. Following \cite{zhou2024mysteryincontextlearningcomprehensive,xie2022explanationincontextlearningimplicit,jiang2023latentspacetheoryemergent}, we generate synthetic language data for training using a doubly-embedded Markov chain model. To simulate varying user prompt ambiguity $\epsilon$, we inject noise into the transition matrix and uniformly distribute prompt ambiguity within $[\epsilon_{\textrm{min}},\epsilon_{\textrm{max}}]$. We scale system factors proportionally to GAI models' utility to analyze their interactions. Due to the page limit, we include additional results, such as those for different ambiguity distributions, in appendix \cite{Online_appendix}, where the insights are similar.

\subsection{The User's Optimal Strategy and Expected Payoff}
\subsubsection{The User's Optimal Strategy}
\begin{figure}[ht]
\vspace{-4pt}
    \centering
    \subfigure[Prompt Number and Ambiguity]{\label{ICL}\includegraphics[width=0.23\textwidth]{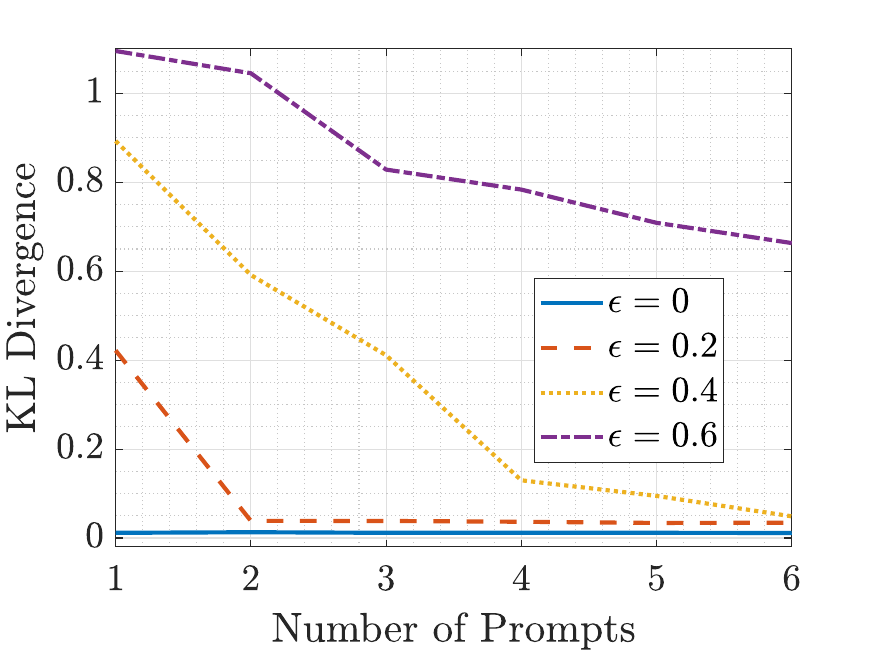}}
    \subfigure[Optimal Number $n^*(m,p_m,\epsilon)$]{\label{op_n}\includegraphics[width=0.23\textwidth]{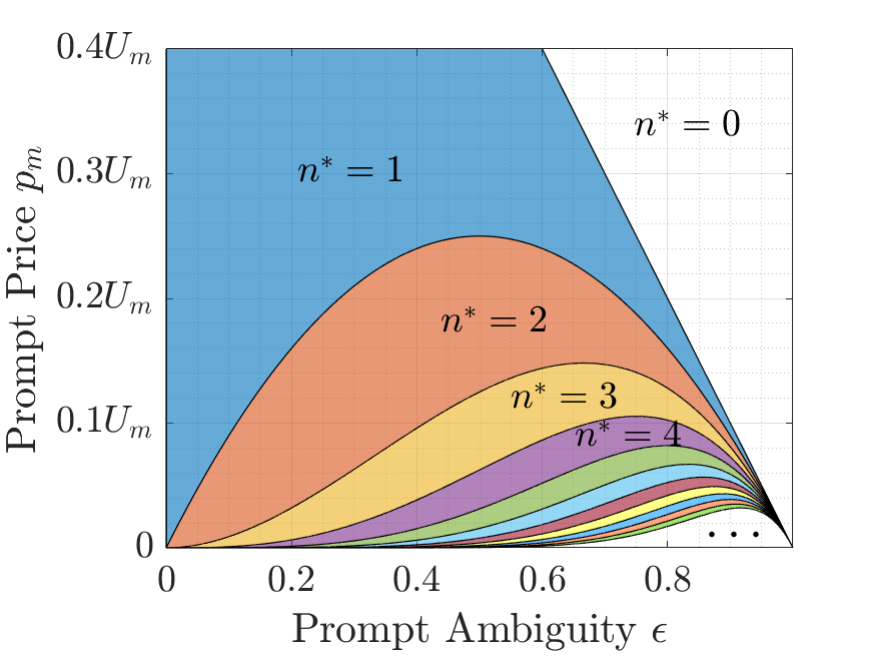}}
    \vspace{-6pt}
    \caption{The User's Optimal Utilization Strategy.}
    \vspace{-2pt}
    \label{F_user1}
\end{figure}

Figure \ref{F_user1} illustrates how prompt ambiguity $\epsilon$ and the number of prompts $n$ influence the GAI model’s output and the user’s optimal strategy. In Figure \ref{ICL}, the KL divergence \cite{KL} between the model’s output and the user’s true intention increases with ambiguity $\epsilon$ but decreases as the number of prompts $n$ grows. When $\epsilon=0$ (no ambiguity), the divergence is minimal, aligning with the analysis in Section \ref{sys2.1}. Figure \ref{op_n} shows that users with moderate ambiguity tend to issue more prompts, while a higher prompt price $p_m$ reduces the number of prompts.
\subsubsection{The User's Optimal Payoff}
\begin{figure}[ht]
\vspace{-4pt}
    \centering
    \subfigure[Payoff $\pi^*_{\textrm{user}}$ and Price $p_{s^*}$]{\label{op_user_e}\includegraphics[width=0.23\textwidth]{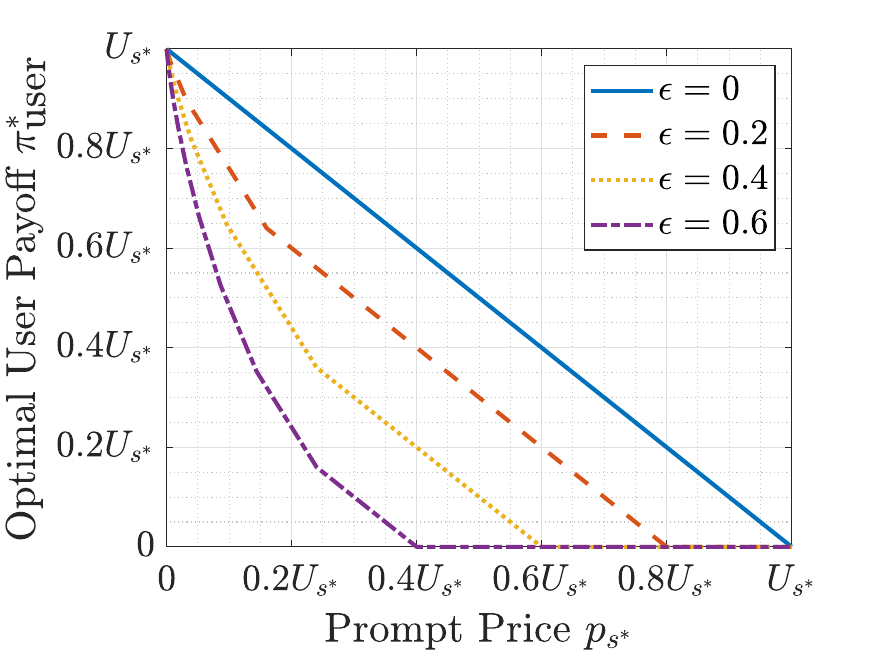}}
    \subfigure[Payoff $\pi^*_{\textrm{user}}$ and Ambiguity $\epsilon$]{\label{op_user_p}\includegraphics[width=0.23\textwidth]{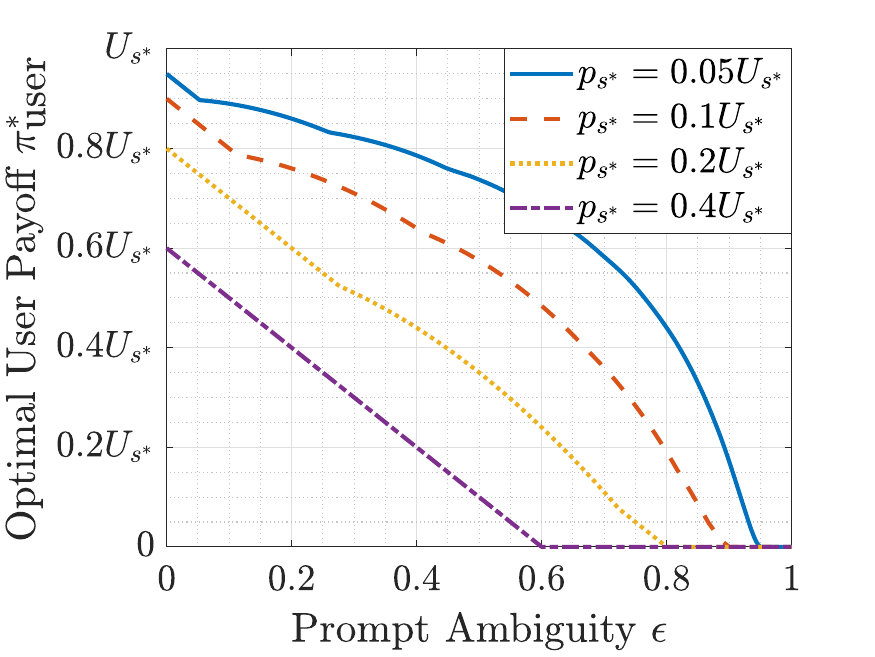}}
    \vspace{-6pt}
    \caption{The User's Optimal Expected Payoff.}
    \vspace{-2pt}
    \label{F_user2}
\end{figure}

Figure \ref{F_user2} demonstrates how the user’s optimal payoff $\pi^*_{\textrm{user}}$, given the model’s optimal decision $s^*$, varies with ambiguity $\epsilon$ and prompt price $p_{s^*}$. As seen in Figure \ref{op_user_e}, ambiguity $\epsilon$ determines the price threshold $p_{s^*}$ at which the user achieves a positive payoff and is willing to purchase the AIGC service. Figure \ref{op_user_p} reveals that as $p_{s^*}$ increases, the user’s payoff $\pi^*_{\textrm{user}}$ declines despite strategic adjustments in utilization. Notably, higher ambiguity $\epsilon$ leads to a sharper reduction in the user’s optimal payoff.
\subsection{The Platform's Optimal Pricing and Expected Payoff} 
\subsubsection{The Platform's Optimal Pricing} 
\begin{figure}[ht]
\vspace{-4pt}
    \centering
    \subfigure[Price $p^*_{m^*}$ and Ambiguity $\epsilon$]{\label{opp_e}\includegraphics[width=0.23\textwidth]{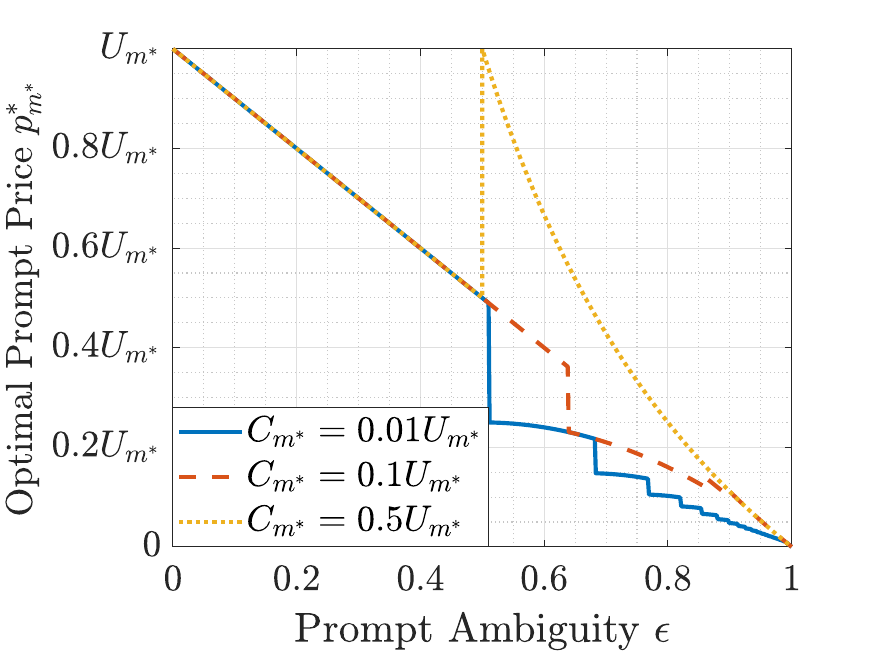}}
    \subfigure[$n^*(m^*,p^*_{m^*},\epsilon)$ with $p^*_{m^*}$]{\label{op_n_un_opp}\includegraphics[width=0.23\textwidth]{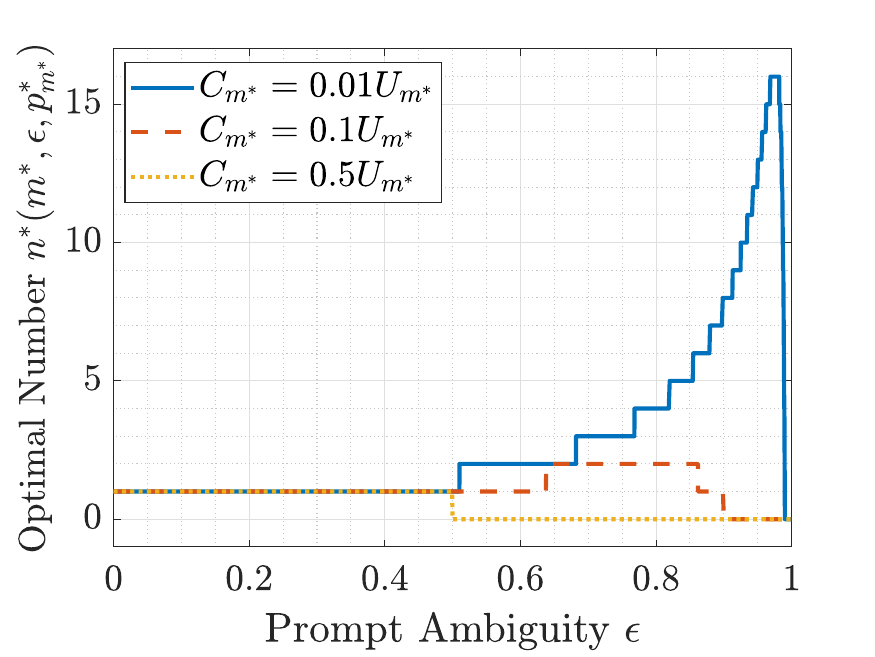}}
    \vspace{-6pt}
    \caption{The Platform's Optimal Prompt Pricing.}
    \vspace{-2pt}
    \label{F_pla1}
\end{figure}

Figure \ref{F_pla1} examines the platform’s optimal prompt price $p^*_{m^*}$ for model $m^*$ under varying prompt ambiguities $\epsilon$ and the resulting user strategy $n^*(m^*, p^*_{m^*}, \epsilon)$. In Figure \ref{opp_e}, the optimal pricing behavior depends on prompt cost $C_{m^*}$ and ambiguity $\epsilon$. Initially, as $\epsilon$ increases, the platform lowers $p^*_{m^*}$ to encourage more prompts and maximize its payoff. However, beyond a critical ambiguity threshold (e.g., $\epsilon = 0.86$ for $C_{m^*} = 0.1U_{m^*}$, marked by the red dashed line in both subfigures), the platform shifts to increasing $p^*_{m^*}$, prioritizing higher prices over more frequent prompts for payoff maximization.
\subsubsection{The Platform's Optimal Payoff} 
\begin{figure}[ht]
\vspace{-4pt}
    \centering
    \subfigure[Setup of $U_H=1.8U_L$, $C_H=0.04U_L$ and $C_L=0.02U_L$]{\label{comp_s1}\includegraphics[width=0.23\textwidth]{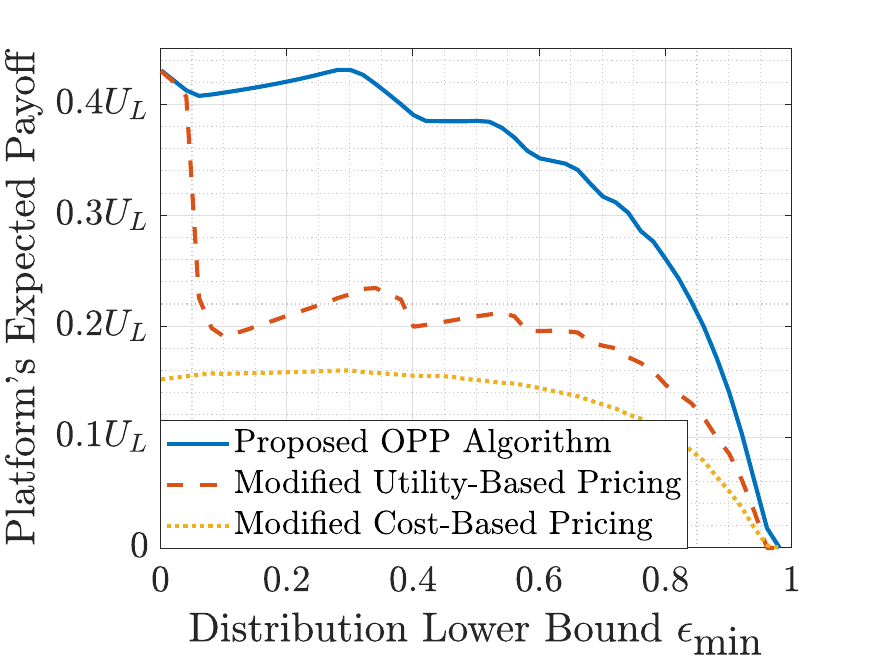}}
    \subfigure[Setup of $U_H=1.5U_L$, $C_H=0.06U_L$ and $C_L=0.02U_L$]{\label{comp_s2}\includegraphics[width=0.23\textwidth]{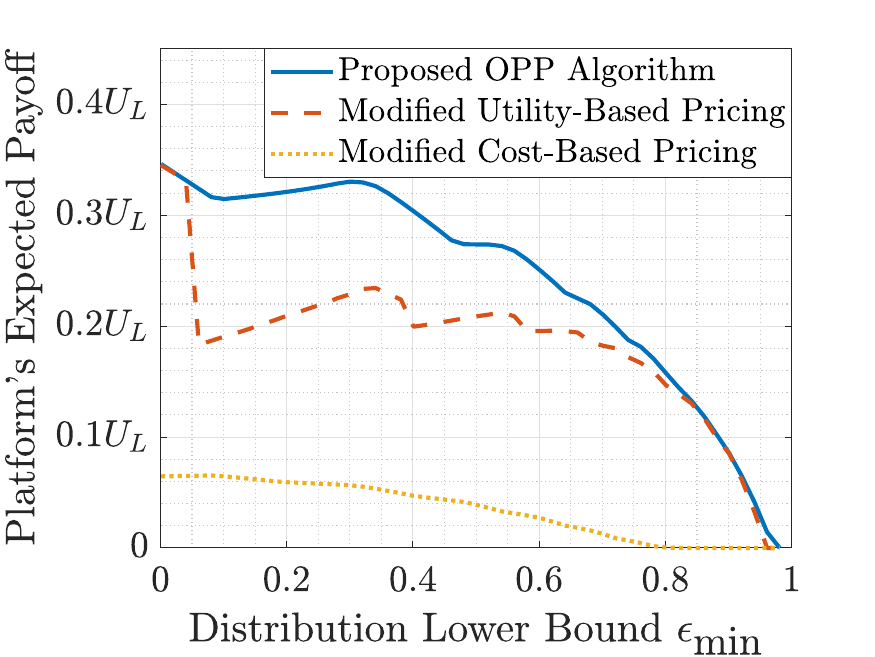}}
    \vspace{-6pt}
    \caption{The Platform's Optimal Expected Payoff ($\epsilon_{\textrm{max}}=1$).}
    \vspace{-4pt}
    \label{F_pla2}
\end{figure}

Figure \ref{F_pla2} compares the platform's payoff under the OPP algorithm for model set $\mathcal{M} = \{M_L, M_H\}$ with two benchmarks:
\begin{itemize}
    \item Modified Utility-Based Pricing \cite{9300226}: Adapts value-based pricing but neglects users’ strategic decisions across model sets.
    \item Modified Cost-Based Pricing \cite{Courcoubetis2007-gv}: Incorporates prompt costs but overlooks heterogeneous user strategies, limiting payoff optimization.
\end{itemize}

As shown in Figure \ref{F_pla2}, the platform’s payoff decreases with lower $\epsilon_{\text{min}}$ under all pricing mechanisms. While both OPP and utility-based pricing exhibit fluctuations due to user strategies, the OPP algorithm consistently outperforms benchmarks by leveraging users’ two-step decisions. It achieves at least $75.05\%$ and $31.72\%$ higher payoffs in the setups of Figures \ref{comp_s1} and \ref{comp_s2}, respectively.
\section{Conclusion}
\label{S7}
This paper provides the first study on strategic interactions between users and the platform in AIGC services. By introducing prompt ambiguity, we derive the optimal user utilization strategies and platform prompt pricing. Our exploration reveals complex interactions among decisions and system factors, where ambiguity adversely affects the payoff of both parties.

In future works, we will refine our theoretical analysis and OPP algorithm to support GAI models with more granular performance tiers, improving adaptability to diverse applications. Meanwhile, we will enhance our experiments with practical data sets and model-based agents.

\section*{Acknowledgment}
This work is supported by National Natural Science Foundation of China (Project 62472367, 62102343, 62271434), Guangdong Research (Project 2021QN02X778), Shenzhen Key Lab of Crowd Intelligence Empowered Low-Carbon Energy Network (No. ZDSYS20220606100601002), Shenzhen Stability Science Program 2023, Shenzhen Institute of Artificial Intelligence and Robotics for Society, Longgang District Shenzhen's ``Ten Action Plan'' for Supporting Innovation Projects (No. LGKCSDPT2024002), Shenzhen Science and Technology Program (Project JCYJ20230807114300001, JCYJ20220818103006012), and Suzhou Frontier Science and Technology Program (Project SYG202310).

\bibliographystyle{IEEEtran}
\bibliography{ref}
\end{document}